\documentclass[sigconf]{acmart}
\usepackage{algorithm}
\usepackage{algorithmic}
\usepackage{graphicx}
\usepackage{subfigure}
\usepackage{caption}
\usepackage{natbib}
\usepackage{bm}
\usepackage{subcaption}
\usepackage{enumitem}
\usepackage{amsthm}

\usepackage{enumitem}
\usepackage{multicol}
\usepackage{balance}

\AtBeginDocument{%
  \providecommand\BibTeX{{%
    \normalfont B\kern-0.5em{\scshape i\kern-0.25em b}\kern-0.8em\TeX}}}

\copyrightyear{2023}
\acmYear{2023}
\setcopyright{acmlicensed}
\acmConference[CIKM '23] {Proceedings of the 32nd ACM International Conference on Information and Knowledge Management}{October 21--25, 2023}{Birmingham, United Kingdom.}
\acmBooktitle{Proceedings of the 32nd ACM International Conference on Information and Knowledge Management (CIKM '23), October 21--25, 2023, Birmingham, United Kingdom}
\acmPrice{15.00}
\acmISBN{979-8-4007-0124-5/23/10}
\acmDOI{10.1145/3583780.3615463}

\begin{document}

\title{Continual Learning in Predictive Autoscaling}


\settopmatter{authorsperrow=4}

\author{Hongyan Hao}
\affiliation{%
  \institution{Ant Group}
  \country{China}
}
\email{hongyanhao.hhy@alipay.com}

\author{Zhixuan Chu}
\affiliation{%
  \institution{Ant Group}
  \country{China}
}
\email{chuzhixuan.czx@alipay.com}

\author{Shiyi Zhu}
\affiliation{%
  \institution{Ant Group}
  \country{China}
}
\email{zhushiyi.zsy@antgroup.com}

\author{Gangwei Jiang}
\affiliation{%
  \institution{Ant Group}
  \country{China}
}
\email{gangwei.jgw@antgroup.com}

\author{Yan Wang}
\affiliation{%
  \institution{Ant Group}
  \country{China}
}
\email{luli.wy@antgroup.com}

\author{Caigao Jiang}
\affiliation{%
  \institution{Ant Group}
  \country{China}
}
\email{caigao.jcg@antgroup.com}

\author{James Y Zhang}

\affiliation{%
  \institution{Ant Group}
  \country{USA}
}
\email{james.z@antgroup.com}

\author{Wei Jiang}
\affiliation{%
  \institution{Ant Group}
  \country{China}
}
\email{shouzhi.jw@antgroup.com}

\author{Siqiao Xue}
\authornote{Corresponding author.}
\affiliation{%
  \institution{Ant Group}
  \country{China}
}
\email{siqiao.xsq@antgroup.com}

\author{Jun Zhou}
\affiliation{%
  \institution{Ant Group}
  \country{China}
}
\email{jun.zhoujun@antgroup.com}

\renewcommand{\shortauthors}{Hongyan Hao et al.}

\begin{abstract}
  Predictive Autoscaling is used to forecast the workloads of servers and prepare the resources in advance to ensure service level objectives (SLOs) in dynamic cloud environments.
  However, in practice, its prediction task often suffers from performance degradation under abnormal traffics caused by external events (such as sales promotional activities and applications' re-configurations), for which a common solution is to re-train the model with data of a long historical period, but at the expense of high computational and storage costs.
  To better address this problem, we propose a replay-based continual learning method, i.e., \textbf{D}ensity-based \textbf{M}emory \textbf{S}election and \textbf{H}int-based Network Learning \textbf{M}odel (DMSHM), using only a small part of the historical log to achieve accurate predictions.
  First, we discover the phenomenon of sample overlap when applying replay-based continual learning in prediction tasks. In order to surmount this challenge and effectively integrate new sample distribution, we propose a density-based sample selection strategy that utilizes kernel density estimation to calculate sample density as a reference to compute sample weight and employs weight sampling to construct a new memory set.
  Then we implement hint-based network learning based on hint representation to optimize the parameters.
  Finally, we conduct experiments on public and industrial datasets to demonstrate that our proposed method outperforms state-of-the-art continual learning methods in terms of memory capacity and prediction accuracy. Furthermore, we demonstrate remarkable practicability of DMSHM in real industrial applications.
  
\end{abstract}





\maketitle

\section{Introduction}

\begin{figure}
    \centering
    \subfigure[]{
    \begin{minipage}[t]{0.45\linewidth}
    \centering
    \includegraphics[width=0.98\textwidth]{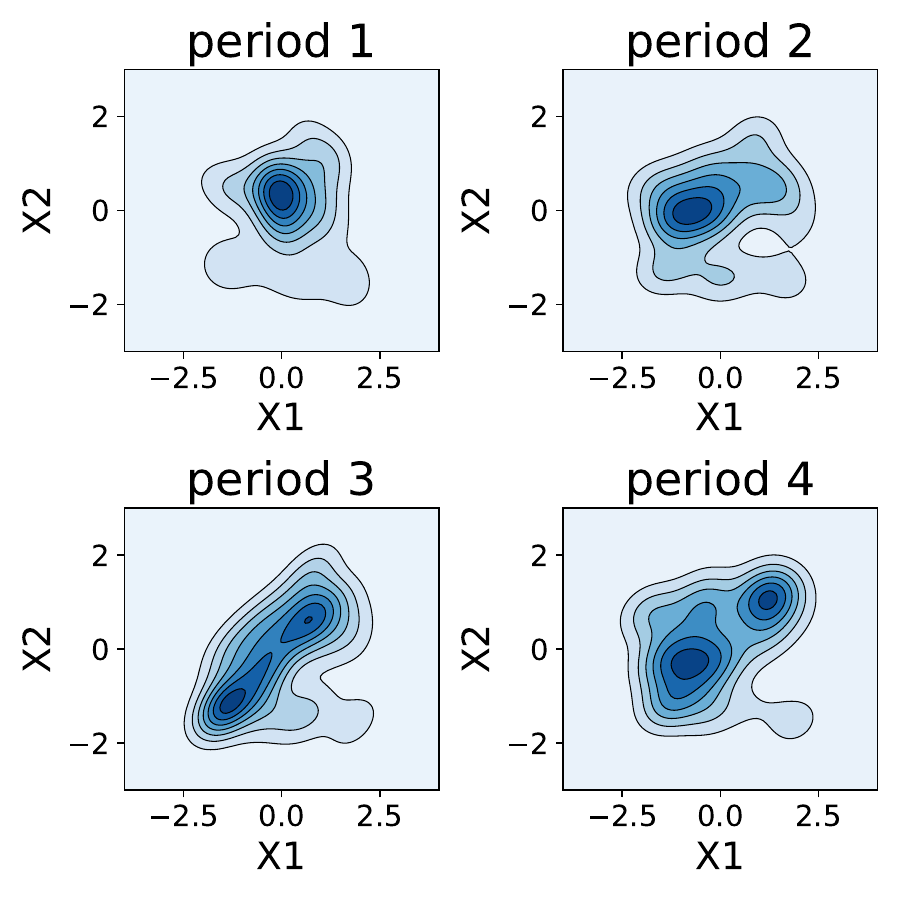}
    \label{fig:sample_dists}
    \end{minipage}
    }
    \subfigure[]{
    \begin{minipage}[t]{0.45\linewidth}
    \centering
    \includegraphics[width=1\textwidth]{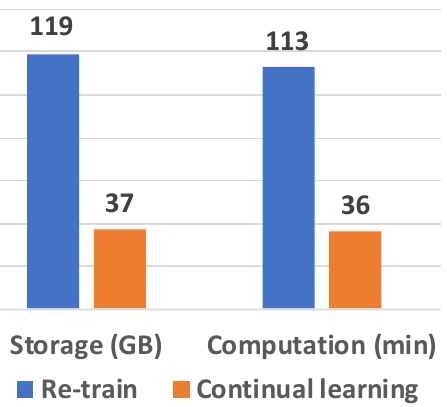}
    \label{fig:resource_comparison}
    \end{minipage}
    }
    \vspace{-3mm}
  \caption{(a) The density distributions of traffic series for 4 consecutive periods, X1 and X2 indicate the first two principle components using PCA. (b) The actual comparison of storage and output time on training data of CPU utilization estimation between re-training and continual learning modes.}
  \label{fig:intros}
\end{figure}

Predictive Autoscaling is a cloud computing method that automatically adjusts cloud services to ensure that resource utilization is maintained within a reasonable range.
This technology has been widely used by major cloud service providers,
such as Google Cloud \cite{google_book} and Azure \cite{azure_book}, helping consumers optimize their resource utilization and cost efficiencies. 
In this paper, we study the optimization of resource usage efficiency on the cloud of Alipay, the world-leading digital payment platform.
The pipeline of Predictive Autoscaling can be divided into 3 steps, i.e., workload forecasting, CPU utilization estimation, and scaling decision. We focus on the first two regression tasks \cite{DBLP:journals/tsc/AbdullahIBPC22,DBLP:conf/kdd/XueQSLZTMWWH0ZL22} in this paper. 

\begin{figure}[h]
  \centering
  \includegraphics[width=\linewidth]{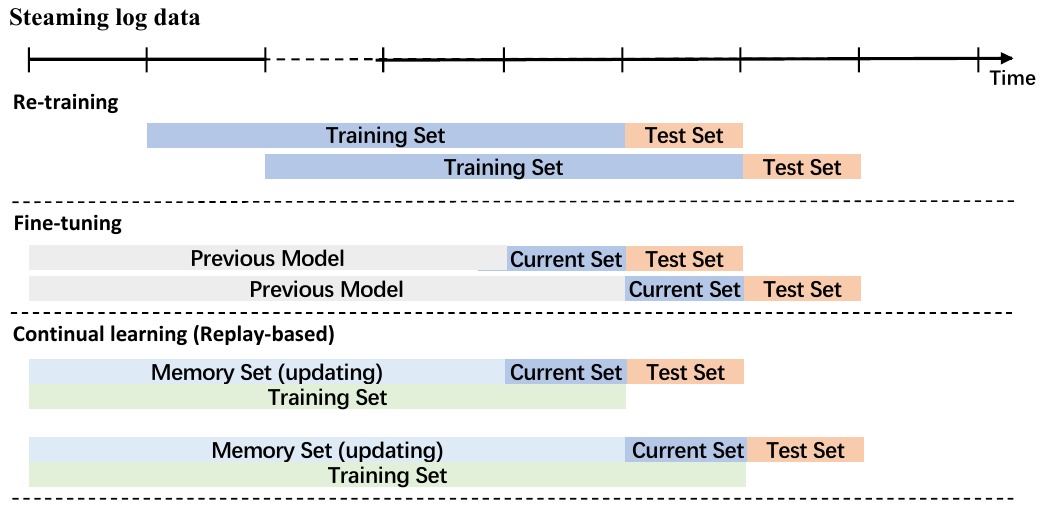}
  \vspace{-3mm}
  \caption{The Predictive Autoscaling pipeline and continual learning mode vs. classic training mode.}
  \Description{The Predictive Autoscaling pipeline}
  \label{fig:training_modes}
\end{figure}

In the above tasks, the data distribution of training samples is usually non-stationary due to external events (such as sales promotional activities and re-configurations of applications), as illustrated in Fig~\ref{fig:sample_dists}, whose data are obtained from an application of Alipay's ecosystem. 
The common solution, as shown in Fig.~\ref{fig:training_modes}, is periodically re-training the model from scratch with a long range of historical data, however, with high costs in terms of both computation and memory consumption.

For example, in order to store a training set of CPU utilization data and to train estimation models for one zone of one application on Alipay's ecosystem, about 119GB of data storage and the computation of 113 minutes of 16000 CPU cores are required. 
One solution is to fine-tune the previously-learned model and make it adaptable to the changes in the data distribution. 
However, this approach may result in catastrophic forgetting, where the new knowledge completely or partially replaces the old knowledge\cite{DBLP:conf/eccv/HayesKSAK20}.

In this paper, we focus on developing a more practical and effective approach for the regression tasks in Predictive Autoscaling.
Inspired by the application of continual learning (CL) \cite{chu2023continual,chu2023continuala}, which learns from a stream of incoming data while avoiding forgetting prior knowledge, we combine replay-based CL with a regression model to solve our problem. As shown in Fig.~\ref{fig:training_modes}, CL maintains a memory set to store informative historical samples and concatenates it with the current data set to serve as the training set. The storage of the memory set is much smaller than that of the single periodic data set. 
In our practical scenario, resource utilization is decreased by almost a quarter of the initial consumption,
as illustrated in Fig.~\ref{fig:resource_comparison}. 
However, after analyzing the properties in the autoscaling scenario, two specific challenges lie in our continual learning paradigm compared with existing works:
i) \textbf{Sample Overlap}: 
In practice,
data distribution of regression problem is usually imbalanced \cite{DBLP:conf/icml/YangZCWK21,DBLP:conf/icml/GongMT22}, 
and besides, sample distributions have certain similarities, leading to area overlapping in distributions between the memory set of replay-based continual learning and the current data set, namely, \textit{sample overlap};
ii) \textbf{Regression Task}: 
 Most continual learning methods are proposed based on the classification tasks, but they cannot be directly applied in a regression task, especially in the process of network training using knowledge distillation \cite{DBLP:journals/spm/ChengWZZ18,DBLP:conf/iccv/SaputraGAMT19}.

To address these challenges, we propose Density-based Memory Selection
and Hint-based network learning Model (DMSHM).
For sample overlap in continual learning, we design a density-based memory selection strategy, utilizing kernel density estimation to calculate sample density as a reference to compute sample weight, followed by constructing a new memory set using weight sampling, to achieve a balance between battling sample overlap and fusing new sample distribution.
To better preserve predicting performance, we apply the hint-based training strategy by storing previous representations and 
utilizing the prior model to produce an intermediate representation of the current sample, which can be regarded as a hint, to recall the prior knowledge.
The ``hint" fills the gap that ``dark knowledge" is inapplicable to regression problems \cite{DBLP:conf/iccv/SaputraGAMT19}. Our main contribution can thus be summarized as follows:
\begin{itemize}[leftmargin=*]
\item We propose a new continual learning method DMSHM for regression tasks in Predictive Autoscaling. We design sample density-based scores to endow with the weight of the sample to construct a memory set and adopt a hint-based network learning strategy to adjust parameters. 
\item We conduct experiments on public and industrial datasets, and demonstrate DMSHM has stronger performance than state-of-the-art continual learning methods on regression tasks.
\end{itemize}

\section{Background}

\subsection{Predictive Autoscaling}

Cloud service providers (CSPs) usually adopt a conservative approach to resource provisioning to satisfy their service level objectives (SLOs). In order to maintain the quality of their services, CSPs often set low CPU utilization targets, even when there are significant variations in the workload \cite{DBLP:conf/cloud/WangZLJRZY0L22}.
This paradigm is inefficient in terms of computation resources, energy consumption, as well as cost, which calls for improvement. 
Predictive Autoscaling forecasts the workload with the help of machine learning models and prepares proper resources in advance to satisfy SLOs with better efficiency \cite{DBLP:journals/tsc/AbdullahIBPC22, DBLP:journals/tsc/SotiriadisBAB19,DBLP:conf/icws/ZhangWPZY20,DBLP:conf/kdd/XueQSLZTMWWH0ZL22,xiao2023automatic}.  
In our industrial practices, we observe that the consumption of storage and computing resources is very high, due to a large number of applications and zones, incentivising further optimization of resource usage and prediction accuracy, for which we devise a new scheme for regression tasks based on continual learning in this paper.

\subsection{Continual Learning}
To ensure the accuracy of our regression model in practical scenarios, a common practice is to periodically retrain using data from the past month. The time span of training dataset is typically chosen so that the cyclical properties of the data can be learned. Continual learning enables us to maintain predictive accuracy on relatively short-term data with reduced resource costs.
However, a common challenge of continual learning is \textit{catastrophic forgetting}, i.e., as new tasks or domains are introduced, the previously acquired knowledge cannot be retained, resulting in performance degradation.
\citeauthor{DBLP:journals/pami/LangeAMPJLST22} \cite{DBLP:journals/pami/LangeAMPJLST22} conduct a critical implementation, compared against mainstream CL methods \cite{DBLP:conf/cvpr/RebuffiKSL17,article,DBLP:conf/iclr/ChaudhryRRE19,DBLP:conf/nips/RolnickASLW19}. A general continual learning (GCL) setting \cite{DBLP:conf/cvpr/AljundiKT19} is also proposed for real-world applications, appending an assumption that the boundaries of tasks are agnostic, based on which, several approaches \cite{DBLP:conf/nips/BuzzegaBPAC20,DBLP:conf/iclr/SunCHLT22} are proposed to make GCL more practical. \citeauthor{DBLP:journals/corr/abs-2101-00926} \cite{DBLP:journals/corr/abs-2101-00926} further propose CLeaR in the context of power forecasting, to address the regression task of GCL. Moreover, the gap of application of knowledge distillation between classification and regression problems are elaborated in \cite{DBLP:journals/spm/ChengWZZ18,DBLP:conf/mipr/TakamotoMI20,DBLP:conf/iccv/SaputraGAMT19}, and to address this problem, we leverage the idea of ``hint'' \cite{DBLP:journals/corr/RomeroBKCGB14} in this work.

\begin{figure*}[htbp]
  \centering
  \includegraphics[width=0.8\linewidth]{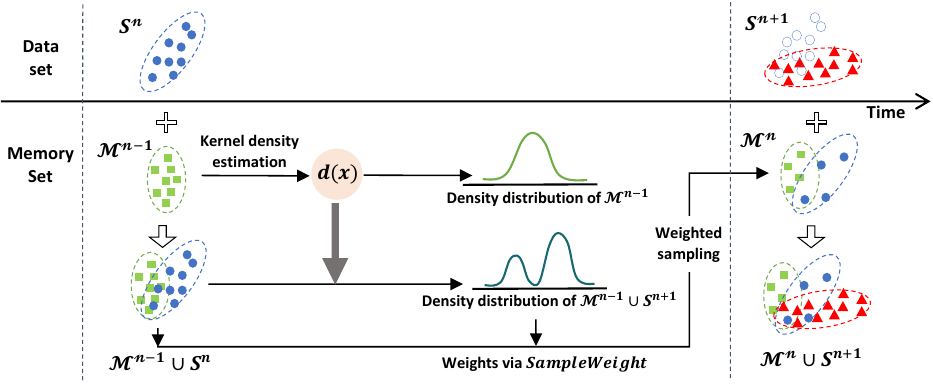}
  \vspace{-3mm}
  \caption{The framework of density-based memory selection. The overall goal is to use weight sampling to select samples from $\mathcal{M}^{n-1}$ and $S^{n}$ as the memory set of the next period. In order to overcome the sample overlap problem, while incorporating the sample in the new distribution into the memory set, we propose two scoring functions \textbf{DensityScore} and \textbf{ShiftLevelScore}.}
  \Description{The framework of density-based memory selection.}
  \label{fig:framework}
\end{figure*}

\section{Preliminary}
As mentioned in the previous section, we try to solve regression tasks (workload forecasting and CPU utilization estimation) in Predictive Autoscaling using continual learning. 
For the convenience of description, we define the $\textit{i}$-th $k-$dimensional input sample of the regression model as $x_{i} \in \mathbb{R}^{k}$ and the label as $y_{i}$, which is a one-dimensional scalar in CPU utilization estimation or a multidimensional vector in workload forecasting. 
In the $\textit{n}$-th period, we acquire dataset $S^{n} = (X^{n}, Y^{n}) = \{ (x_{i}, y_{i}) \}_{i=1}^{N^{n}}$, whose size is $N^{n}$, and the cumulative sample size before present period is $A^{n}=\sum_{i=1}^{n-1}N^{i}$.
In addition, we maintain a memory set $\mathcal{M}^{n}$ with fixed size $M$, which is updated every time a new dataset arrives. 
Using the setting of an ordinary supervised learning problem, we define the representation function as $h^{n}$, indicating an underlying mapping from the input sample to a lower dimension vector $z = h^{n}(x)$. We then define $g^{n}$ to map the representation vector to prediction result. We formulate the prediction function as $f^{n}(x) = g^{n}(h^{n}(x)) = g^{n}(z)$,
whose training parameters are $\theta_h$ and $\theta_g$ for the representation and linear function, respectively. We define the whole parameter set as $\theta$, and it is optimized by defined loss functions which are specified in Section~\ref{sec:param_train} later.

\section{Method}

In the following sections, we detail density-based memory selection and hint-based network learning to overcome catastrophic forgetting from the perspectives of sampling and training, respectively. The former aims to select the most informative samples to represent history to help the model recall the prior knowledge, while the latter uses the hint-based training strategy to train the model with a memory set and current dataset.

\subsection{Density-based Memory Selection}
\label{sec:dms}

Many replay-based methods use reservoir sampling to update the memory set from $\mathcal{M}^{n-1}$ and $S^{n}$, which ensures that each sample has the same probability of being selected into a new memory set $\mathcal{M}^{n}$.  However, data imbalance can result in some data drawn from the overlapping area of the distribution being sampled more than once. Our expectation for the memory set is that it can represent the distribution of historical samples, but the repetition in sample sets can lead to a more severe imbalance, which would impair the efficiency of the memory set.
To overcome this problem, we design a density-based memory selection (DMS) mechanism. 

As shown in Fig.~\ref{fig:framework}, our overall goal is to use weight sampling \cite{zhou2023ptse} to select samples from $\mathcal{M}^{n-1}$ and $S^{n}$ as the memory set of the next period, so we focus on the design of weights. In order to overcome the problem of sample overlap while incorporating the sample in the new distribution into the memory set, we designed two scoring functions, \textbf{DensityScore} and \textbf{ShiftLevelScore}. 

We devise some indicators based on sample density to determine whether the overlapping phenomenon or new distribution occurs. 
Specifically, considering the similarity among the sample distributions across different time steps, 
a smaller mean value difference of $\mathcal{M}^{n-1}$ and $S^{n}$ would result in a smaller variance of combined sample distribution, resulting in the overlap phenomenon.
In principle, the distribution of $\mathcal{M}^{n-1}$ can better represent the real distribution of the previous sample. 
We use kernel density estimation to fit the samples in $\mathcal{M}^{n-1}$, and obtain the mapping function $d(x) \in \mathbb{R}$, formulated as follows:
\begin{equation}
\begin{split}
    d(x) &= \frac{1}{|\mathcal{M}^{n-1}|b^{k}} \sum_{i=1}^{|\mathcal{M}^{n-1}|} K(\frac{x - x_{i}}{b}), \\ 
    K(z) &\geq 0, \quad \int K(z)\mathbf{d}z=1. \nonumber
\end{split}
\label{equation:probs_memory_set}
\end{equation}
We choose the Gaussian kernel function as $K(z)$, and we use maximum likelihood cross-validation to obtain the optimal bandwidth $b$.
Then we use $d(x)$ to calculate the density of each sample in $\mathcal{M}^{n-1} \cup S^{n}$, and samples with very small density are called \textit{border samples}. 
To avoid extreme cases of weight skewness caused by differences in density scales, we employ the $sigmoid$ function to smooth sample density as $q^{n}(x)=1/(1 + e^{-d(x)})$.
We define $q^{n}(x): \mathbb{R}^{k} \to \mathbb{R}_{+}$ as \textbf{DensityScore} function, which outputs a score to reflect the degree of sample aggregation.

\begin{algorithm}[t]
        \caption{DMS \textbf{SampleWeight}}
        \label{alg:sample_score}
        \begin{algorithmic}[1]
        \STATE \textbf{Input:} The memory set $\mathcal{M}^{n-1}$ with budget size $M$;
        The dataset $\mathrm{S}^{n}$ with size $N^{n}$; The cumulative sample size $A^{n}=\sum_{i=1}^{n-1}N^{i}$;
        The \textbf{DensityScore} function $q^{n}$;
        The balance factor $\gamma$.
        \FOR{$i = 1, \dots, (|\mathcal{M}^{n-1}|+N^{n}) $}
            \IF{$i \leq |\mathcal{M}^{n-1}|$}
                \STATE $ (x_{i},z_{i}, y_{i}) \leftarrow  \mathcal{M}^{n-1}_{i}$ \qquad\qquad \text{// $i$-th sample in $\mathcal{M}^{n-1}$}
                \STATE $w(x) = (1-\gamma) \times q^{n}(x_{i}) + \gamma \times \frac{A^{n}}{A^{n}+N^{n}}$
            \ELSE
                \STATE $ (x_{i},y_{i}) \leftarrow  \mathcal{S}^{n}_{i-|\mathcal{M}^{n-1}|}$ \qquad \text{// ($i$-$|\mathcal{M}^{n-1}|$)-th sample in ${S}^{n}$}
                \STATE $w(x) = (1-\gamma) \times q^{n}(x_{i}) + \gamma \times \frac{M}{A^{n}+N^{n}}$
            \ENDIF
        \ENDFOR
        \STATE $ W^{n} := \{w(x)| x \in \mathcal{M}^{n-1} \cup S^{n}\}$
        \RETURN The $n$-th sample weight set $W^{n}$
        \end{algorithmic}
\end{algorithm}

\begin{algorithm}[htp]
        \caption{Density-based Memory Selection}
        \label{alg:dms}
        \begin{algorithmic}[1]
        \STATE \textbf{Input:} The memory set $\mathcal{M}^{n-1}$ with the budget size $M$;
        The dataset $\mathrm{S}^{n}$ with size $N^{n}$; The cumulative sample size $A^{n}=\sum_{i=1}^{n-1}N^{i}$;
        The representation function $ h^{n}(x) $.

        \IF{n = 1}
            \STATE $W^{n} = $ \textbf{SampleWeight}($ \varnothing, M, S^{n}, N^{n}, 0, 0.5, 1$)
            \STATE $(X^{n}_{m}, Y^{n}_{m}) \leftarrow$ \textbf{WeightedSampling}$( S^{n}, W^{n} )$
        \ELSE
            \STATE $d(x) \leftarrow $ \textbf{KernelDensityEstimator}($\mathcal{M}^{n-1}$)
            \STATE $q^{n}(x) = \frac{1}{1 + e^{-d(x)}}$ \qquad\qquad\qquad\text{//} \textbf{DensityScore}
            \STATE $\gamma = $ \textbf{ShiftLevelScore} $( \{ q^{n}(x) | x \in \mathcal{M}^{n-1} \cup S^{n} \})$
            \STATE $W^{n} = $ \textbf{SampleWeight}($ \mathcal{M}^{n-1}, M, S^{n}, N^{n}, A^{n}, q^{n}, \gamma$)
            \STATE $(X^{n}_{m}, Y^{n}_{m}) \leftarrow$ \textbf{WeightedSampling}$( \mathcal{M}^{n-1} \cup S^{n}, W^{n} )$
        \ENDIF
        \STATE $Z^{n}_{m} = \{ z | z = h^{n}(x), x \in X^{n}_{m} \} $ 
        \STATE Update $\mathcal{M}^{n} \leftarrow (X^{n}_{m}, Z^{n}_{m}, Y^{n}_{m})$
        \RETURN The $n$-th memory set $\mathcal{M}^{n}$
        \end{algorithmic}
\end{algorithm}

From the design of the score above, we can infer that the score of samples in the distribution of $\mathcal{M}^{n-1}$ is larger than the rest.  A large number of border samples indicates that the distribution of the combined sample set is quite different from the previous ones, which is called \textit{distribution shift}.  However, sample selection solely relying on the density score can lead to excessive tendency to the samples of $\mathcal{M}^{n-1}$, voiding our attempt to incorporate the new distribution. To mitigate the problem, we devise \textbf{ShiftLevelScore} function that generates the indicator $\gamma$, to describe the level of distribution shift. We use the Gaussian mixture model with two components to fit the density scores of $\mathcal{M}^{n-1} \cup S^{n}$, and we formulate the absolute difference between the means of the two components as $\gamma$. A large $\gamma$ implies the occurrence of distribution shift, and border samples need to be attended, for which 
we introduce the reservoir sampling strategy.
For the batch data stream scenario, we employ two biased coefficients $\frac{A^{n}}{A^{n}+N^{n}}$ and $\frac{M}{A^{n}+N^{n}}$ on samples of $\mathcal{M}^{n-1}$ and $S^{n}$ to ensure that all samples are sampled with equal probability (see Sec.~\ref{sec:bias}).
To trade-off between avoiding distribution shift and mitigating overlap, the coefficient $\gamma$ is used to control reservoir sampling and score-based sampling. We define the procedure as \textbf{SampleWeight} function, which is described in Algorithm~\ref{alg:sample_score}.

After the above processing, we can obtain the sample weight set $W^{n}$, which is then used to select $M$ samples as $\mathcal{M}^{n}$. The detailed description of DMS is shown in Algorithm~\ref{alg:dms}.
Note that the memory set stores not only the selected samples and labels but also the outputs of intermediate representation from the current model, whose intention is explained in Sec.~\ref{sec:param_train} later.

\subsection{Proof of Biased Coefficients}
\label{sec:bias}
The objective of the reservoir sampling algorithm is to choose a fixed number of samples without replacement in the face of an uncertain total number of samples in a single pass while ensuring that each sample is selected with equal probability.
Unlike the traditional reservoir sampling method that draws one sample per period, we process a sample set $S^{n}$ containing $N^{n}$ samples in each period.
The sample set $\mathcal{M}^{n}$ consists of subsets of $\mathcal{M}^{n-1}$ and $S^{n}$ by sampling. Since the samples in $\mathcal{M}^{n-1}$ are sourced from the sample set prior to the $n$-th and $S^{n}$ is being sampled for the first time, it is necessary to assign sampling weights $w^{bias}(x|x \in \mathcal{M}^{n-1})=\frac{A^{n}}{A^{n}+N^{n}}$ and $w^{bias}(x|x \in S^{n})=\frac{M}{A^{n}+N^{n}}$ to the samples in $\mathcal{M}^{n-1}$ and $S^{n}$ to ensure that all samples $ \bigcup_{i=1}^{n}S^{i}$ are sampled with equal probability.  We term the above sampling weights as \textit{biased coefficients}, whose rationality is demonstrated in the mathematical proof below. 

\begin{proof}
The mathematical induction can be divided into the initial step and the inductive step.

\begin{itemize}[leftmargin=*]
    \item \textbf{Initial step}: For $n=1$, $A^{1}=0$ and $|\mathcal{M}^{0}|=0$, then
    \begin{align}
        w^{bias}(x|x \in \mathcal{M}^{0}) &=\frac{A^{1}}{A^{1}+N^{1}}=\frac{0}{0+N^{1}}=0 \nonumber \\
        w^{bias}(x|x \in S^{1}) &=\frac{M}{A^{1}+N^{1}}=\frac{M}{0+N^{1}}=\frac{M}{N^{1}}. \nonumber \\
        w(x|x \in S^{1}) &= w^{bias}(x|x \in S^{1}) = \frac{M}{A^{1}+N^{1}}  \nonumber
    \end{align}
    As of the $\text{1-st}$ period's end, all samples $ \bigcup_{i=1}^{1}S^{i}$ are sampled with equal probability.
    \item \textbf{Inductive step}:
    We assume that the proposition is true for $n=k$, i.e., after applying biased coefficients $w^{bias}(x|x \in \mathcal{M}^{k-1})=\frac{A^{k}}{A^{k}+N^{k}}$ and $w^{bias}(x|x \in S^{k})=\frac{M}{A^{k}+N^{k}}$, all samples $\bigcup_{i=1}^{k}S^{i}$ have the same weights:
    \begin{align}
        w(x|x \in \mathcal{M}^{k-1})=w(x|x \in S^{k})=\frac{M}{A^{k}+N^{k}} . \nonumber
    \end{align}
    For $n=k+1$, after applying biased coefficients, the sample weight of $\mathcal{M}^{k}$ and $S^{k+1}$ are
    \begin{align}
        w(x|x \in \mathcal{M}^{(k+1)-1}) &= w(x|x \in \mathcal{M}^{k-1}) \times w^{bias}(x|x \in \mathcal{M}^{(k+1)-1}) \nonumber \\ 
        &= \frac{M}{A^{k}+N^{k}} \times \frac{A^{k+1}}{A^{k+1}+N^{k+1}} \nonumber \\ 
        &= \frac{M}{A^{k+1}+N^{k+1}}, \nonumber
    \end{align}
    and 
    \begin{align}
        w(x|x \in {S}^{k+1}) &= \frac{M}{A^{k+1}+N^{k+1}}, \nonumber
    \end{align}
    i.e., the same weight of $\mathcal{M}^{k}$ and $S^{k+1}$ means the samples of $\bigcup_{i=1}^{k+1}S^{i}$ have the same probability of being sampled.
\end{itemize}
\end{proof}

\subsection{Hint-based Network Learning}
\label{sec:param_train}
To optimize parameters for mitigating catastrophic forgetting, most approaches leverage dark knowledge \cite{DBLP:journals/corr/HintonVD15} to retain the prior knowledge, where the main idea is to store softened logits output, which is then used to guide the optimization trajectory of the current model. However, this approach does not apply to the regression problem. Specifically, the output of the regression task is a continuous value, which has the same properties as the ground truth along with an unknown error distribution \cite{DBLP:conf/mipr/TakamotoMI20,DBLP:journals/spm/ChengWZZ18}, so that keeping the previous dark knowledge of samples is not a suitable solution for our problem.
Inspired by the theory of \citeauthor{DBLP:journals/corr/RomeroBKCGB14} \cite{DBLP:journals/corr/RomeroBKCGB14}, which suggests that intermediate representation provides ``hint'' for the current model to imitate the previous one, we devise our hint-based network learning strategy.

During the network training, we have memory set $\mathcal{M}^{n-1}$ and current dataset $S^{n}$. Before training in the current step, we utilize the representation network of the previous step $h^{n-1}(x)$ to generate additional hint $Z^{n}_{hint} = h^{n-1}(X^{n})$. 
We define the loss $\mathcal{L}_{hint}$, which is the error of intermediate output $\hat{Z}^{n}_{hint} = h^{n}(X^{n})$ and $Z^{n}_{hint}$, formulated as:
\begin{equation}
    \mathcal{L}_{hint} := \mathbb{E}_{(x,y) \sim S^{n}} \left[ l(h^{n-1} (x), h^{n} (x) ) \right] ,
    \nonumber
\end{equation}
where $l(h^{n-1} (x), h^{n} (x) )$ is the mean absolute error between $\hat{Z}^{n}_{hint}$ and $Z^{n}_{hint}$, and $l$ means the same for the rest of the paper.
The network $h^{n}$ regards $Z^{n}_{hint}$ as guidance for conducting training close to the previous optimization trajectory, which assists the network in recalling knowledge to alleviate catastrophic forgetting.

We also minimize the error between hints $Z^{n-1}_{m}$ and intermediate representation $\hat{Z}^{n-1}_{m} = h^{n}(X^{n-1}_{m})$ of memory samples, along with the error between their ground truth $Y^{n-1}_{m}$ and predictions $\hat{Y}^{n-1}_{m} = f^{n}(X^{n-1}_{m})$, and we refer to the loss as $\mathcal{L}_{memory}$:
\begin{equation}
\begin{split}
    \mathcal{L}_{memory} &:= \alpha 
\cdot \mathbb{E}_{(x,z,y) \sim \mathcal{M}^{n-1}} \left[ l(z, h^{n}(x)) \right] \nonumber \\
    &\quad \ + \beta \cdot 
 \mathbb{E}_{(x,z,y) \sim \mathcal{M}^{n-1}} \left[ l(y, f^{n}(x)) \right], \nonumber 
 \end{split}
\end{equation}
where the coefficients $\alpha$ and $\beta$ serve as weights to adjust the importance of following the optimization trajectory of the former model ($\theta^{n-1}$) and recalling from ground-truth labels $Y^{n-1}_{m}$. 

Furthermore, the model needs to acquire new knowledge from the current dataset $S^{n}=(X^{n}, Y^{n})$, and we set the network to minimize weighted mean absolute error, defined as:
\begin{equation}
    \mathcal{L}_{cur} := \mathbb{E}_{(x,y) \sim S^{n}} \left[ l(y, f^{n} (x) ) \right] ,
    \nonumber
\end{equation}
where the $\mathcal{L}_{cur}$ reflect the convergence process on new data set. 

To integrate the three losses $\mathcal{L}_{hint}$, $\mathcal{L}_{memory}$ and $\mathcal{L}_{cur}$, we sum the terms up with coefficients $\delta$ and $\xi$ to trade-off tendencies between recalling previous information and learning new knowledge, to form the total loss function ($\mathcal{L}$) as follows:
\begin{equation}
    \mathcal{L} = \mathcal{L}_{memory} + \xi \cdot \mathcal{L}_{hint} + \delta \cdot \mathcal{L}_{cur} .
    \nonumber
\end{equation}

\section{Experiments}
\subsection{Datasets} 
To evaluate the effectiveness of our proposed method, we conduct experiments on ATEC's\footnote{https://github.com/TRaaSStack/Forecasting} public industrial dataset, which is for workload forecasting competition.  We select one traffic series and split all data into $8$ periodic datasets in chronological order to perform the evaluation of the workload forecasting task. Furthermore, we also perform the evaluation of CPU utilization estimation (CUE) on a zone of a main application whose data was collected from an industrial scenario. 
It includes $9$ types of traffic values and CPU usage per minute of 8 days, whose distributions are significantly different from each other, as shown in Fig.~\ref{fig:cpu_dist}. We need to model the mapping between traffic and CPU usage.

\subsection{Baselines and Metrics}
Fine-tuning \cite{DBLP:journals/pami/LangeAMPJLST22} is a naive baseline for our comparison of prediction performance and we further compare against DER++ \cite{DBLP:conf/nips/BuzzegaBPAC20} and CLeaR \cite{DBLP:journals/corr/abs-2101-00926}. The former is an excellent general continual learning method, and the latter applies continual learning on regression tasks. 
We use different networks as $h(x)$ for two regression tasks, i.e., an LSTM layer for ATEC task and an MLP layer for CUE task. 

To measure memorization and forecast performance of the model in a continual learning setup, we regard the first dataset as a historical set and the samples composed of random selection from all $8$ datasets as a future set.
After every period, we record the mean square error (MSE) of the model on the historical and future set. 
We define forgetting error (FE) as the average MSE of all steps on the historical set, which indicates the ability to memorize and overcome catastrophic forgetting.
In addition, we define the prediction error (PE) as the MSE of the final period on the future set, reflecting the performance of model generalization. A small PE value means better predictive power for samples from unknown distributions.

\subsection{Experiment Results}

\begin{table*}
  \vspace{-2mm}
  \caption{Mean and standard deviation (in bracket) of experiment results on ATEC and CUE datasets.}
  \vspace{-2mm}
  \label{tab:experiments}
  \begin{tabular}{ccccl}
    \toprule
    & \multicolumn{2}{c}{ATEC} & \multicolumn{2}{c}{CUE} \\
    \cmidrule(lr){2-3} \cmidrule(lr){4-5}
    Models & FE & PE & FE & PE\\
    \midrule
Finetune & 0.0686 (0.0312) & 0.0739 (0.0437) & 33.46 (5.39) & 23.59 (6.29) \\ 
CLeaR    & 0.0512 (0.0245) & 0.0405 (0.0276) & 20.57 (4.62) & 19.02 (4.38) \\ 
DER++    & 0.0450 (0.0367) & 0.0315 (0.0125) & 18.12 (2.84) & 17.05 (2.93) \\ 
\midrule
DMSHM (w/o DMS)    & 0.0445 (0.0129) & 0.0311 (0.0174) & 17.91 (2.83) & 16.99 (2.72) \\ 
DMSHM (w/o Hint)    & 0.0426 (0.0201) & 0.0301 (0.0214) & 16.82 (1.92) & 16.90 (2.84) \\ 
\textbf{DMSHM} (ours)    & \textbf{0.0404 (0.0167)} & \textbf{0.0282 (0.0186)} & \textbf{15.70 (2.14)} & \textbf{16.85 (3.46)} \\ 
  \bottomrule 	 
\end{tabular}
\end{table*}

\begin{figure}[htbp]
  \centering
  \subfigure[]{
  \begin{minipage}[t]{0.45\linewidth}
    \centering
    \includegraphics[width=1\textwidth]{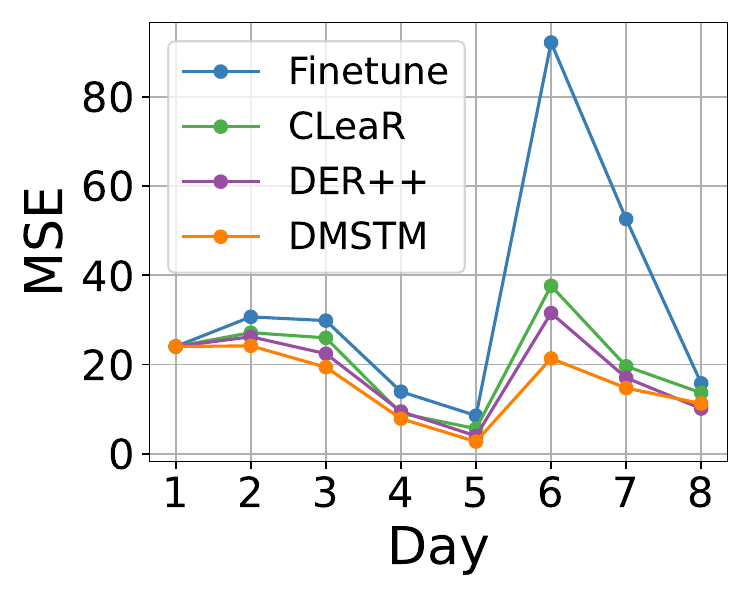}
    \vspace{-8mm}
    \label{fig:mse_history_point}
  \end{minipage}
  }
  \subfigure[]{
  \begin{minipage}[t]{0.45\linewidth}
    \centering
    \includegraphics[width=1\textwidth]{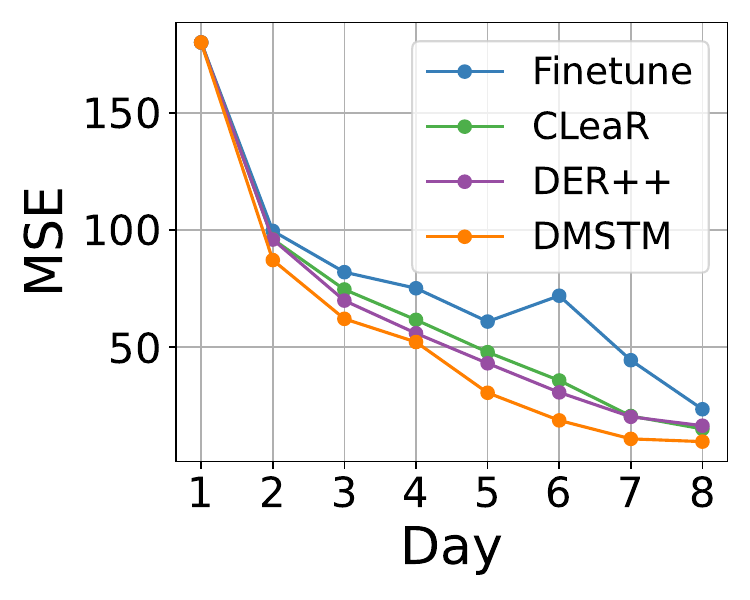}
    \vspace{-8mm}
    \label{fig:mse_future_point}
  \end{minipage}
  }
   \vspace{-2mm}
  \subfigure[]{
  \begin{minipage}[t]{0.6\linewidth}
    \centering
    \includegraphics[width=1\textwidth]{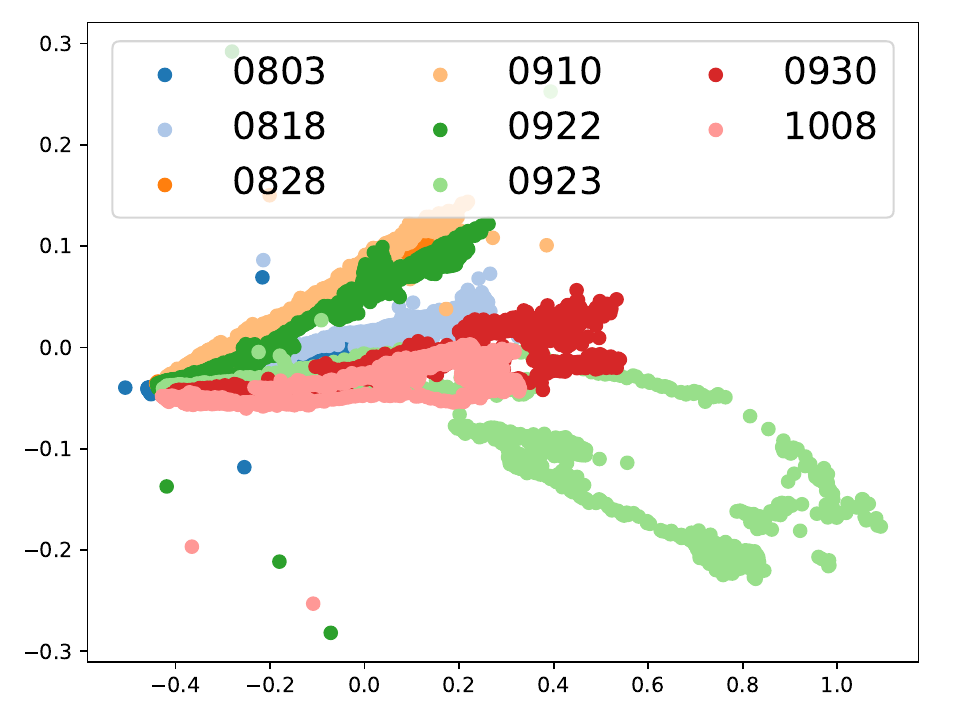}
    \vspace{-2mm}
    \label{fig:cpu_dist}
  \end{minipage}
  }
  \vspace{-4mm}
  \caption{The MSE of $4$ models on historical set (a) and future set (b) of CUE. (c) is the scatter plot of data sets from all $8$ days, using PCA to reduce dimensions.}
  \vspace{-5mm}
\end{figure}

The experiment results are shown in Table~\ref{tab:experiments}, where the best results are highlighted in bold, where DMSHM obtains the best performance in terms of forgetting error and prediction error on both ATEC dataset and CPU utilization estimation task.
Finetune acquires the worst performance on all metrics due to the absence of capability to overcome catastrophic forgetting. To explore the performance of the model in each period, we demonstrate MSE of the four models on the historical set and future set in 8 days, as illustrated in Fig.~\ref{fig:mse_history_point} and Fig.~\ref{fig:mse_future_point}. The MSE on the historical set of all models appears to have a tendency to decrease before the fifth day due to the fact that the datasets of these days are of high similarity.  However, on the sixth day, when a configuration change takes place, the sample distribution has a somewhat larger change, as shown in Fig.~\ref{fig:cpu_dist}. 
DMSHM is the most robust to this disturbance due to the density-based selection strategy. As for the prediction error, DMSHM shows smaller MSEs, which indicates that the samples in the memory set are more similar to the overall distribution, and the model is capable of predicting data in the future set.

\subsection{Ablation Study}

In order to evaluate the effectiveness of the density-based memory selection strategy (DMS) and hint-based network learning method (Hint) of the proposed DMSHM model, we conduct two ablation studies of DMSHM(w/o DMS) and DMSHM(w/o Hint) on all datasets.  Our first study is DMSHM(w/o DMS), where the density-based memory selection method is replaced by a reservoir sampling strategy, which assigns equal weight to all samples in the memory set and current data set in \textbf{SampleWeight} function. Given the practical sample storage scheme of batch acquisition for a new dataset, we assign weights $\frac{A^{n}}{A^{n}+N^{n}}$ and $\frac{M}{A^{n}+N^{n}}$ to the samples in memory set and the new dataset to obtain uniform selection probability for each sample.
The second ablation study is DMSHM(w/o Hint), where the hint-based network learning is replaced by conventional parameter adjustment strategy with loss function $\mathcal{L}_{memory} + \delta \cdot \mathcal{L}_{cur}$, which invalidates the model's capability of maintaining historical knowledge through hint-based parameter learning. 

According to the results provided in Table~\ref{tab:experiments}, the performance worsens after removing either DMS or the hint-based module compared to the original DMSHM.  Specifically, after replacing the DMS with a reservoir sampling strategy, DMSHM(w/o DMS) suffers a dramatic decline in performance.  Besides, DMSHM(w/o DMS) exhibits a more drastic decrease, which indicates DMS is more critical for addressing catastrophic forgetting and maintaining the forecasting accuracy, compared to the hint parameter learning.

\section{Deployment}

Our method has been successfully deployed in Alipay Cloud to serve over 20,000 zones across more than 3,000 applications, as evidenced by its impressive effectiveness and efficiency. In one representative zone, the data time span is reduced from 113 minutes to just 36 minutes, while the storage size decreased from 119 GB to 37 GB. Moreover, our method achieves CPU utilization estimation with only one-third of the original resource usage, as illustrated in Fig.~\ref{fig:resource_comparison}.
As the dataset size increases, the degree of resource-saving needs to be calculated in conjunction with the size of the current data set and memory set that replaces a large training set. The existence of the memory set exponentially reduces the amount of data needed to be stored. By significantly reducing the capital cost of application operation, our method makes a valuable contribution towards the environmental sustainability of Alipay.

\section{Conclusion}
The focus of this paper is to address the problem of inefficient resource utilization in regression tasks within Predictive Autoscaling. To solve this challenge, we propose a novel continual learning approach called DMSHM. This method is designed to efficiently manage sample overlap through a density-based memory selection strategy that leverages reservoir sampling with batch stream samples. To support the rationality of two biased coefficients, we have included inductive mathematical proof. In addition, we have implemented a hint-based network learning strategy to bridge the gap between regression tasks and continual learning. Our model has been extensively tested, and the results demonstrate its effectiveness.

\clearpage
\bibliographystyle{ACM-Reference-Format}
\balance
\bibliography{sample-base}










\end{document}